\documentclass[lettersize,journal]{IEEEtran}

\usepackage{amsmath,amsfonts}
\usepackage{algorithmicx}
\usepackage{algorithm}
\usepackage{array}
\usepackage[caption=false,font=normalsize,labelfont=sf,textfont=sf]{subfig}
\usepackage{textcomp}
\usepackage{stfloats}
\usepackage{url}
\usepackage{verbatim}
\usepackage{graphicx}
\usepackage{cite}
\usepackage{multirow}
\usepackage{diagbox}
\usepackage{pifont}
\usepackage{algpseudocode}
\usepackage{xcolor}
\usepackage{placeins}

\usepackage{amssymb}
\usepackage{hyperref}
\hypersetup{
    colorlinks=true,
    linkcolor=black,     
    citecolor=blue,     
    urlcolor=black       
}
\usepackage{booktabs}
\usepackage{makecell}

\begin{document}

\title{Beyond Euclidean Prototypes: Spectral Disentanglement and Geodesic Matching for Few-Shot Medical Image Segmentation}

\author{Penghao Jia, Zhiyong Huang*, Mingyang Hou, Zhi Yu, Shuai Miao, Jiahong Wang, Yan Yan

\thanks{*Corresponding author: Zhiyong Huang.
Penghao Jia, Zhiyong Huang, Mingyang Hou, Zhi Yu, Shuai Miao are with the School of Microelectronics and Communication Engineering, Chongqing University, Chongqing 400044, China (e-mail: zyhuang@cqu.edu.cn);
Jiahong Wang is with the Health Management Institute, The Second Medical Center, and the National Clinical Research Center for Geriatric Diseases, Chinese PLA General Hospital, Beijing 100853, China;
Yan Yan is with Shenzhen Institutes of Advanced Technology, Chinese Academy of Sciences, Shenzhen 518055, China (e-mail: yan.yan@siat.ac.cn).
}
}



\maketitle

\begin{abstract}
Few-Shot Medical Image Segmentation (FSMIS) aims to delineate novel anatomical targets from one or a few annotated support images, addressing the annotation scarcity in medical imaging. Notwithstanding recent advancements, current prototype-based methods are bottlenecked by two coupled limitations: 1) cue entanglement, where a single spatial-domain prototype is forced to summarise organ silhouette, parenchymal texture and boundary appearance simultaneously, so any support–query mismatch on one cue propagates indiscriminately to the others; and 2) topology-blind matching, where cosine similarity measures distance in the ambient Euclidean space and ignores the connectivity of the underlying feature manifold, causing fragmented activations inside low-contrast organs and leakage into neighbouring tissues. To this end, we propose Spectral--Geodesic Prototype Network (SGP-Net), built around a Spectral--Geodesic Prototype Module with two coupled components. A Spectral Prototype Bank (SPB) decomposes support and query features into low-, mid- and high-frequency bands via learnable radial Fourier filters, yielding three disentangled prototypes per class that separately encode shape, texture and boundary cues. A Geodesic Matcher (GM) then replaces cosine similarity with a differentiable heat-diffusion approximation of geodesic distance, propagating matching signals along a feature affinity graph so that on-manifold pixels accumulate consistent responses while off-manifold look-alikes are suppressed. Experiments on three public FSMIS benchmarks demonstrate that SGP-Net achieves competitive performance against recent state-of-the-art methods. The code is available at https://github.com/naivejph/SGP-Net.git.
\end{abstract}


\section{Introduction}
\IEEEPARstart{M}{edical} image segmentation \cite{tcsvt1,tcsvt3,r47} underpins a wide range of clinical workflows, from disease diagnosis to treatment planning and longitudinal follow-up \cite{r1}. Modern deep networks \cite{r2,r3,r44,r45} have pushed the accuracy of automated segmentation to new heights, yet their training appetite is at odds with the realities of medical practice: pixel-wise annotation is expensive, demands clinical expertise, and is further constrained by patient privacy and the long tail of rare diseases \cite{r4}. How to deliver accurate segmentation under such severe annotation scarcity has therefore become a pressing question for the community.

An intuitive solution is provided by few-shot learning \cite{tcsvt4,tcsvt5,magenet}, which mimics the human ability to grasp a new concept from a handful of examples. Cast in the episodic meta-learning framework, Few-Shot Medical Image Segmentation (FSMIS) trains a model on base classes and asks it to delineate novel anatomical targets at test time given only one or a few annotated support slices, with no further fine-tuning \cite{r5,r6}. Among the various technical routes, prototype-based methods have become the de-facto standard: a class-representative vector is summarised from the support image via masked average pooling and matched against query features by cosine similarity \cite{r7,r8}. Building on this template, recent works have enriched prototypes with multi-vector decomposition \cite{r9}, superpixel-based self-supervision \cite{r10}, uncertainty-aware filtering \cite{r11}, multi-descriptor generation \cite{r12}, and dual-interspersion designs \cite{r13}, yielding steady gains across standard benchmarks.

\begin{figure}[t]
\centering
\includegraphics[width=\linewidth]{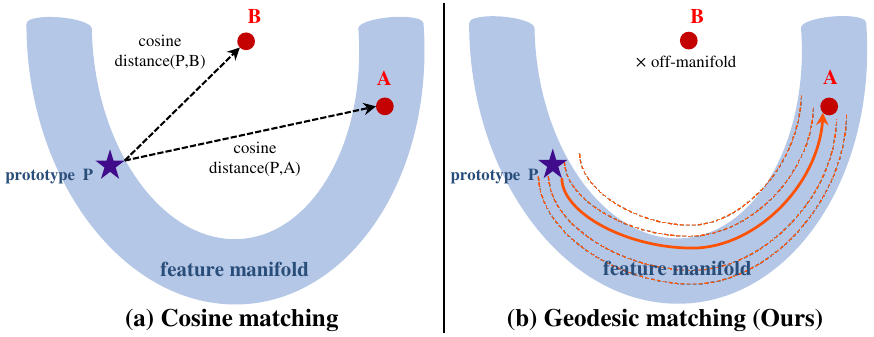}
\caption{Limitation of cosine matching. (a) Cosine similarity takes a Euclidean shortcut and ranks off-manifold pixel $B$ closer to prototype $P$ than the on-manifold pixel $A$. (b) Our Geodesic Matcher diffuses signals along the manifold, correctly favouring $A$ over $B$.}
\label{fig:gm_0}
\end{figure}

Despite the progress achieved thus far, we observe that current FSMIS pipelines remain bottlenecked by two issues that are largely orthogonal to prototype quantity. (i) Cue entanglement in the support representation. A single pooled vector is forced to encapsulate organ silhouette, parenchymal texture, and boundary appearance all at once. Once these heterogeneous cues are aggregated into one channel space, any mismatch between support and query on a single cue propagates indiscriminately to the others during matching. Multi-prototype variants \cite{r6} partition the entangled vector across spatial sub-regions, but the entanglement itself is never resolved. (ii) Topology-blind matching. As illustrated in Fig.~\hyperref[fig:gm_0]{\ref*{fig:gm_0}(a)}, cosine similarity measures distance in the ambient Euclidean feature space and is therefore prone to two symmetric vulnerabilities: a query pixel A that lies on the same feature manifold as the prototype P but is geodesically distant gets a low score, while an off-manifold pixel B that happens to be Euclidean-close to P—e.g. a neighbouring organ with similar intensity statistics—gets a spuriously high one. These problems translate directly into fragmented activations inside low-contrast organs and leakage into surrounding tissues, precisely the regions where segmentation boundaries are most sensitive to error\cite{maccfa,r15}.

We argue that resolving these two issues calls for changes on both ends of the matching pipeline: the prototype side needs to separate the cues it carries, and the matching side needs to respect the geometry along which features actually live. Following this view, we propose a Spectral--Geodesic Prototype Network (SGP-Net) for FSMIS, built around a Spectral--Geodesic Prototype Module with two coupled components. A Spectral Prototype Bank (SPB) sends features through a learnable radial decomposition in the 2-D Fourier domain and produces one prototype per frequency band, so that shape, texture, and boundary information are carried by different channels rather than fused into one. A Geodesic Matcher (GM) then closes the matching loop with a heat-method approximation of geodesic distance: as sketched in Fig.~\hyperref[fig:gm_0]{\ref*{fig:gm_0}(b)}, matching signals are diffused along a feature affinity graph, so that pixels lying on the same manifold as P (such as A) accumulate strong responses while off-manifold pixels (such as B) decay. The two modules run twice per episode under a foreground–background symmetric design with all parameters shared. Extensive experiments on three benchmark datasets demonstrate the superiority of SGP-Net. Our contributions are summarised as follows:

\begin{itemize}
\item We identify two coupled limitations of current FSMIS prototype methods: cue entanglement, where a single vector is forced to carry organ silhouette, texture and boundary cues simultaneously, and topology-blind matching, where cosine similarity ignores the connectivity of the underlying feature manifold. Both are addressed within a unified Spectral--Geodesic Prototype Module.
\item We propose the Geodesic Matcher, a differentiable heat-diffusion module that lifts prototype–query matching from ambient Euclidean space onto the feature manifold, suppressing off-manifold look-alikes while restoring connectivity within low-contrast organs.
\item We propose the Spectral Prototype Bank, which factorises features into low-, mid- and high-frequency bands via learnable radial Fourier filters and produces one disentangled prototype per band. Unlike prior frequency-aware FSMIS methods that operate at the feature level, our decomposition is performed at the prototype level.
\end{itemize}

\section{Related Works}
\subsection{Few-Shot Semantic Segmentation}
Few-Shot Semantic Segmentation (FSS) \cite{r42,r43} adapts few-shot learning for pixel-wise prediction, aiming to identify novel classes using minimal annotated support images without requiring further fine-tuning. The field originated with Shaban et al. \cite{r16}, who introduced a two-branch conditional framework, while PANet \cite{r8} later established the widely used prototypical baseline via non-parametric prototype matching and an alignment regularizer. Since then, research has generally followed two main trajectories. On the prototype-based front, models quickly evolved past single-vector representations: Liu et al. \cite{r17} extracted part-aware prototypes from the foreground and Li et al. \cite{r19} utilized superpixel clustering for adaptive allocation. Feature enhancement also saw developments, with PFENet \cite{r9} incorporating high-level similarity priors into query features, and Lang et al. \cite{r20} leveraging base-class knowledge from background regions to explicitly define what to exclude. Taking a complementary approach, dense matching methods focus on correlation. HSNet \cite{r21} aggregates these multi-level connections using 4-D convolutions, MIANet \cite{r23} effectively combines both routes by integrating instance- and category-level cues. Despite these structural advancements, nearly all these frameworks share a critical underlying assumption: they rely on standard spatial metrics like cosine or dot products to calculate prototype–query similarity. While this approach is effective for natural images, it proves fundamentally fragile when applied to the low-contrast and highly homogeneous characteristics typical of medical imaging.

\subsection{Few-Shot Medical Image Segmentation}
The success of FSS has motivated its extension to medical imaging, where annotation scarcity is acute. Existing FSMIS methods \cite{r4,r39,famnet} fall into two streams. Interaction-based methods couple the support and query branches through dedicated feature interaction modules. SE-Net \cite{r25} pioneers this design with squeeze-and-excitation gating, GCN-DE \cite{r26} models support–query relations through a global correlation network, CRAPNet \cite{r27} enforces bidirectional consistency via cycle-resemblance attention, and PGRNet \cite{r28} casts the interaction as prototype-guided graph reasoning.

Prototype-based methods, which are closer to our work, account for most recent progress. SSL-ALPNet \cite{r10} pools local prototypes from superpixel-derived pseudo-labels, while ADNet \cite{r29} and ADNet++ \cite{r11} cast segmentation as anomaly detection anchored by a single foreground prototype. Q-Net \cite{r15} learns query-informed adaptive thresholds, RPT \cite{r30} performs region enhancement with a transformer, and both GMRD \cite{r12} and DSPNet \cite{r6} pursue more representative descriptors. More recent works continue to refine prototype generation: PAMI \cite{r31} partitions the support foreground into sub-regions and debiases the resulting prototypes, UPRE-Net \cite{r4} fuses global and local prototypes under uncertainty-guided filtering, SPENet \cite{r32} adapts the number of local prototypes to lesion size, and DIFD \cite{r13} generates basis prototypes through dual interspersion and deploys them flexibly across query positions.

A small number of works have begun to explore the frequency domain. FAMNet \cite{famnet} performs frequency-aware feature alignment to bridge cross-domain distribution gaps, treating the spectrum as a domain-invariance regulariser. MCCFA \cite{maccfa} introduces frequency-aware enhancement inside its cross-attention module, where spectral cues modulate feature responses but a single entangled prototype is still extracted afterwards. In contrast, our SPB performs decomposition at the prototype level, so that low-, mid- and high-frequency cues are carried by separate class representatives all the way into the matching stage. Beyond this, none of the above methods replaces Euclidean matching with a manifold-aware metric, which is precisely the gap our Geodesic Matcher fills.

\section{Methodology}

\subsection{Problem Definition}
We follow the standard episodic protocol of FSMIS. The training set $\mathcal{D}_{\mathrm{train}}$ and the test set $\mathcal{D}_{\mathrm{test}}$ are drawn from disjoint label spaces $\mathcal{C}_{\mathrm{train}}$ and $\mathcal{C}_{\mathrm{test}}$ with $\mathcal{C}_{\mathrm{train}}\cap\mathcal{C}_{\mathrm{test}}=\varnothing$, so that target organs at test time are unseen during training. Each episode is sampled as a 1-way $K$-shot task that contains a support set $\mathcal{S}=\{(I_s^k, M_s^k)\}_{k=1}^{K}$ and a query $\mathcal{Q}=\{(I_q, M_q)\}$, where $I_s^k, I_q \in \mathbb{R}^{H \times W}$ are the support and query slices and $M_s^k, M_q \in \{0,1\}^{H \times W}$ are their binary masks of the target class. The model is trained on $\mathcal{D}_{\mathrm{train}}$ to predict $\hat{M}_q$ from $(I_q, \mathcal{S})$ without further fine-tuning on $\mathcal{D}_{\mathrm{test}}$. We instantiate the 1-way 1-shot setting in this work and report results on novel classes from $\mathcal{D}_{\mathrm{test}}$.

\begin{figure*}[t]
\centering
\includegraphics[width=\linewidth, height=0.5\textheight, keepaspectratio]{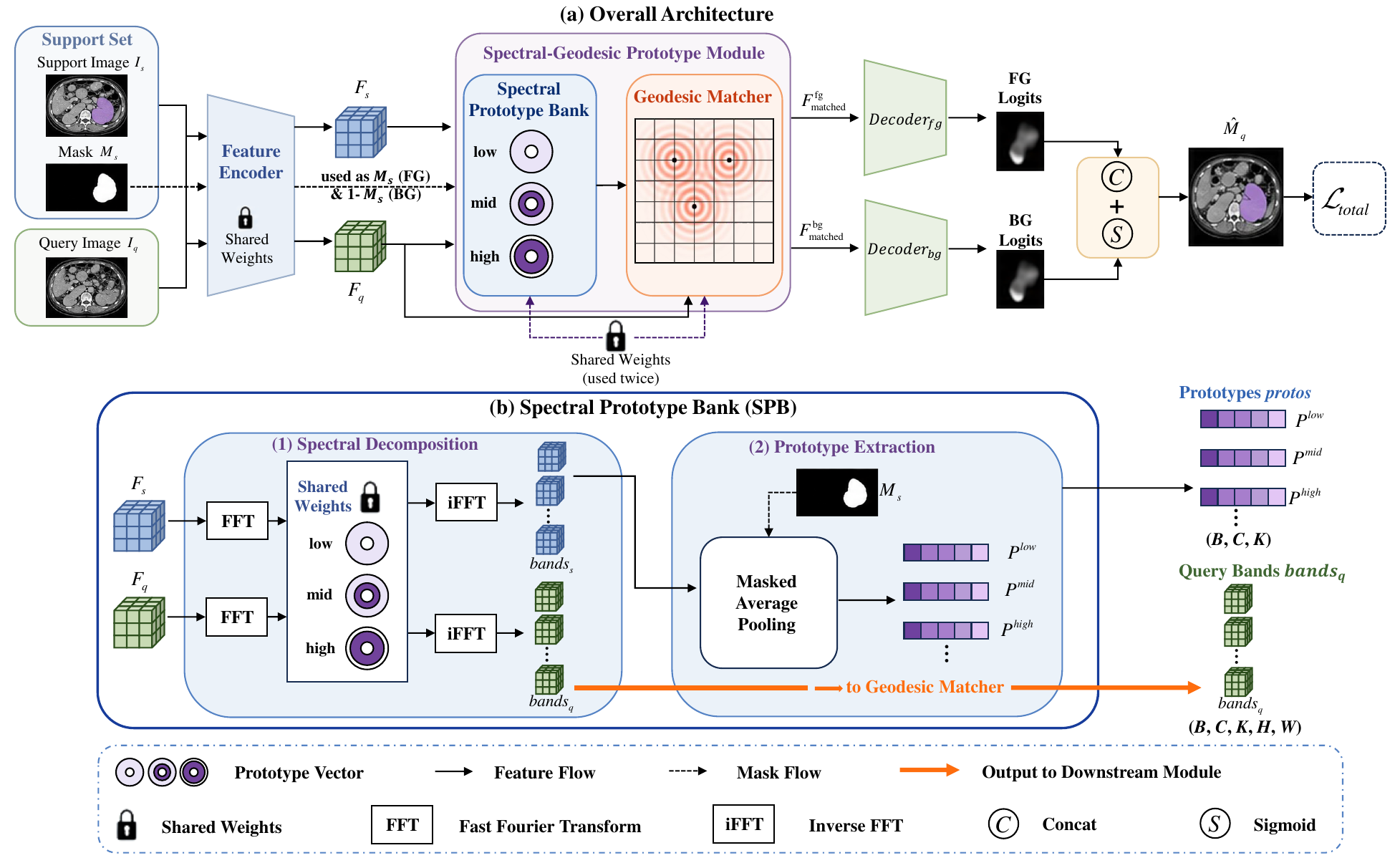}
\caption{Overview of the proposed SGP-Net. \textbf{(a) Overall architecture.} The Spectral--Geodesic Prototype Module, composed of a Spectral Prototype Bank (SPB) and a Geodesic Matcher (GM), is invoked twice with shared parameters: once with the support mask $M_s$ for the foreground branch and once with $1-M_s$ for the background branch, driving two parallel decoders to produce the final prediction. \textbf{(b) Spectral Prototype Bank.} SPB decomposes $F_s$ and $F_q$ into three learnable radial frequency bands via FFT, and extracts one prototype $P^{(k)}$ per band by Masked Average Pooling.}
\label{fig:overview}
\end{figure*}

\subsection{Architecture Overview}

Fig.~\hyperref[fig:overview]{\ref*{fig:overview}(a)} shows the architecture of SGP-Net. A weight-shared feature encoder $f_{\theta}(\cdot)$ first maps the support image $I_s$ and the query image $I_q$ to deep feature maps $F_s, F_q \in \mathbb{R}^{B \times C \times h \times w}$. These features, together with the support mask $M_s$, are then fed into the proposed Spectral--Geodesic Prototype Module, which is composed of a Spectral Prototype Bank (SPB) and a Geodesic Matcher (GM). The SPB decomposes $F_s$ and $F_q$ in the 2-D Fourier domain into $K=3$ learnable radial bands and produces one prototype $P^{(k)}$ per band, yielding the band features $bands_q$ and the prototype set $\mathcal{P}$. The GM then matches every query pixel to each band prototype with a heat-diffusion-based geodesic similarity and blends the three prototypes per pixel, producing a Decoder-compatible tensor $F_{\mathrm{matched}} \in \mathbb{R}^{B \times (2C+3) \times h \times w}$. The whole module is invoked twice per episode, once with $M_s$ as the foreground mask and once with $1-M_s$ as the background mask, with all SPB and GM parameters shared between the two calls. The two outputs $F_{\mathrm{matched}}^{\mathrm{fg}}$ and $F_{\mathrm{matched}}^{\mathrm{bg}}$ drive two parallel decoders $\mathrm{Decoder}_{\mathrm{fg}}$ and $\mathrm{Decoder}_{\mathrm{bg}}$ that emit the foreground and background logits, which are concatenated and passed through a softmax to obtain the final prediction $\hat{M}_q$. The whole network is trained end-to-end by minimising $\mathcal{L}_{\mathrm{total}}$.

\subsection{Spectral Prototype Bank}
\label{subsec:spb}
Most FSS methods \cite{r4,r6,r40,r41} represent each support class with one or several prototypes obtained in the spatial domain via masked average pooling, so a single $C$-dimensional vector simultaneously encodes organ silhouette, parenchymal texture, and boundary information. Any support--query inconsistency on one cue therefore propagates to the others during matching, and multi-prototype variants only redistribute this entanglement over smaller spatial regions. We therefore propose a Spectral Prototype Bank (SPB) that decomposes the support and query features in the 2-D Fourier domain into $K=3$ learnable radial bands and produces one prototype per band, so that the three cues are represented separately rather than jointly.

A shared encoder $f_{\theta}(\cdot)$ first extracts the support and query feature maps, $F_s = f_{\theta}(I_s)$ and $F_q = f_{\theta}(I_q) \in \mathbb{R}^{B\times C\times h\times w}$. As illustrated in Fig.~\hyperref[fig:overview]{\ref*{fig:overview}(b)}, SPB then proceeds with Spectral Decomposition followed by Prototype Extraction.

\subsubsection{Spectral Decomposition}
We map both $F_s$ and $F_q$ into the Fourier domain with a 2-D real-input FFT under orthonormal scaling,
\begin{align}
\widetilde{F}_s = \mathcal{F}\{F_s\}, \quad \widetilde{F}_q = \mathcal{F}\{F_q\},
\end{align}
where $\widetilde{F}_s, \widetilde{F}_q \in \mathbb{C}^{B \times C \times h \times (\lfloor w/2 \rfloor + 1)}$ and the half-spectrum fully represents real-valued feature maps via conjugate symmetry. We then partition the half-spectrum along its radial frequency $\rho(u,v) = \sqrt{\nu_y(u)^2 + \nu_x(v)^2}$, where $\nu_y, \nu_x$ are the discrete normalised frequencies returned by \texttt{fftfreq} and \texttt{rfftfreq}. Two learnable cut-off radii $r_1 < r_2$ split the radial axis into low, mid, and high bands. To keep them strictly positive and strictly increasing, we parameterise the gaps rather than the radii directly,
\begin{align} \label{eq:2}
r_1 = \mathrm{softplus}(\tilde{r}_1), \quad r_2 = r_1 + \mathrm{softplus}(\tilde{r}_2^{\mathrm{gap}}),
\end{align}
initialised so that $r_1 \approx 0.25$ and $r_2 \approx 0.55$. To preserve gradient flow through the radii, hard indicators are replaced by smooth sigmoid roll-offs with sharpness $\beta = \mathrm{softplus}(\tilde{\beta}) + 1$,
\begin{equation} \label{eq:3}
\begin{aligned}
\mathcal{M}^{\mathrm{low}}(\rho)  &= \sigma\big(\beta(r_1 - \rho)\big), \\
\mathcal{M}^{\mathrm{mid}}(\rho)  &= \sigma\big(\beta(r_2 - \rho)\big) - \sigma\big(\beta(r_1 - \rho)\big), \\
\mathcal{M}^{\mathrm{high}}(\rho) &= 1 - \sigma\big(\beta(r_2 - \rho)\big),
\end{aligned}
\end{equation}
where $\sigma(\cdot)$ is the logistic sigmoid. By construction, $\mathcal{M}^{\mathrm{low}}(\rho)+\mathcal{M}^{\mathrm{mid}}(\rho)+\mathcal{M}^{\mathrm{high}}(\rho)=1$ for all $\rho$, so the spectral energy of each Fourier coefficient is split, not duplicated, across the three bands.

Applying each band mask in the Fourier domain and inverting the transform yields the band-restricted feature maps,
\begin{align} \label{eq:4}
F_s^{(k)} = \mathcal{F}^{-1}\!\big\{\mathcal{M}^{(k)} \odot \widetilde{F}_s\big\}, \quad
F_q^{(k)} = \mathcal{F}^{-1}\!\big\{\mathcal{M}^{(k)} \odot \widetilde{F}_q\big\},
\end{align}
for $k \in \{\mathrm{low},\mathrm{mid},\mathrm{high}\}$, where $\mathcal{F}^{-1}\{\cdot\}$ is the orthonormal irFFT2. We collect them as $bands_s = [F_s^{(\mathrm{low})}, F_s^{(\mathrm{mid})}, F_s^{(\mathrm{high})}]$ and $bands_q$ analogously, with $bands_s, bands_q \in \mathbb{R}^{B \times C \times K \times h \times w}$.

\subsubsection{Prototype Extraction}
We generate one prototype per band via Masked Average Pooling. The support mask $M_s \in [0,1]^{B \times H \times W}$ is first downsampled to the encoder resolution as $M_s^{\downarrow} = \mathrm{Interp}(M_s, (h, w))$. The $k$-th band prototype is then
\begin{align} \label{eq:5}
P^{(k)} = \frac{\sum_{(i,j)} F_s^{(k)}(i, j) \odot M_s^{\downarrow}(i, j)}{\max\!\big(\sum_{(i,j)} M_s^{\downarrow}(i, j),\,\varepsilon\big)},
\end{align}
where $P^{(k)} \in \mathbb{R}^{B \times C}$ and $\varepsilon = 10^{-5}$ guards against division by zero. The three prototypes are stacked into $\mathcal{P} = [P^{\mathrm{low}}, P^{\mathrm{mid}}, P^{\mathrm{high}}] \in \mathbb{R}^{B \times C \times K}$. The entire SPB is invoked twice per episode, once with $M_s$ and once with $1 - M_s$, with all parameters shared. The output $(bands_q, \mathcal{P})$ is then forwarded to the Geodesic Matcher.

\subsubsection{Frequency--Anatomy Correspondence}
A pertinent question is \emph{why} a radial decomposition of the feature spectrum should align with anatomical semantics. We give a brief justification that links the band partition $(r_1, r_2)$ to the physical scale of organ structures.

By the Wiener--Khinchin theorem, the energy of a feature channel at radial frequency $\rho$ equals the Fourier transform of its spatial autocorrelation, so the energy at $\rho$ is concentrated on autocorrelation lags $\ell \propto 1/\rho$. The radial partition therefore maps directly to three spatial-scale regimes. Low frequencies ($\rho < r_1$) correspond to lags $\ell \gtrsim 1/r_1$, comparable to the size of an entire organ; this regime captures the slowly varying envelope that distinguishes the organ from its surroundings, namely the \emph{global silhouette}. Mid frequencies ($r_1 \le \rho < r_2$) correspond to intermediate lags on the order of intra-organ sub-structures such as vessels and lobules, so they carry \emph{parenchymal texture}. High frequencies ($\rho \ge r_2$) correspond to short lags of a few pixels, dominated in medical scans by the intensity step at organ contours, so they concentrate on \emph{boundary cues}.

Two further points reinforce this correspondence. First, the $1/\rho$ relation between radial frequency and spatial scale is independent of the specific organ, modality or encoder, which explains why the same band partition transfers across Abd-MRI, Abd-CT and CMR. Second, the role of the two learnable radii $(r_1, r_2)$ is only to localise the dataset-specific boundaries between these three regimes; the mapping itself is determined by the spatial-scale argument above and not by the optimisation. The convergence trajectory of $(r_1, r_2)$ in Fig.~\ref{fig:radii_curve} and the per-band activation maps in Fig.~\ref{fig:band_act} provide direct empirical support for this analysis.

\begin{figure*}[t]
\centering
\includegraphics[width=\linewidth, height=0.3\textheight, keepaspectratio]{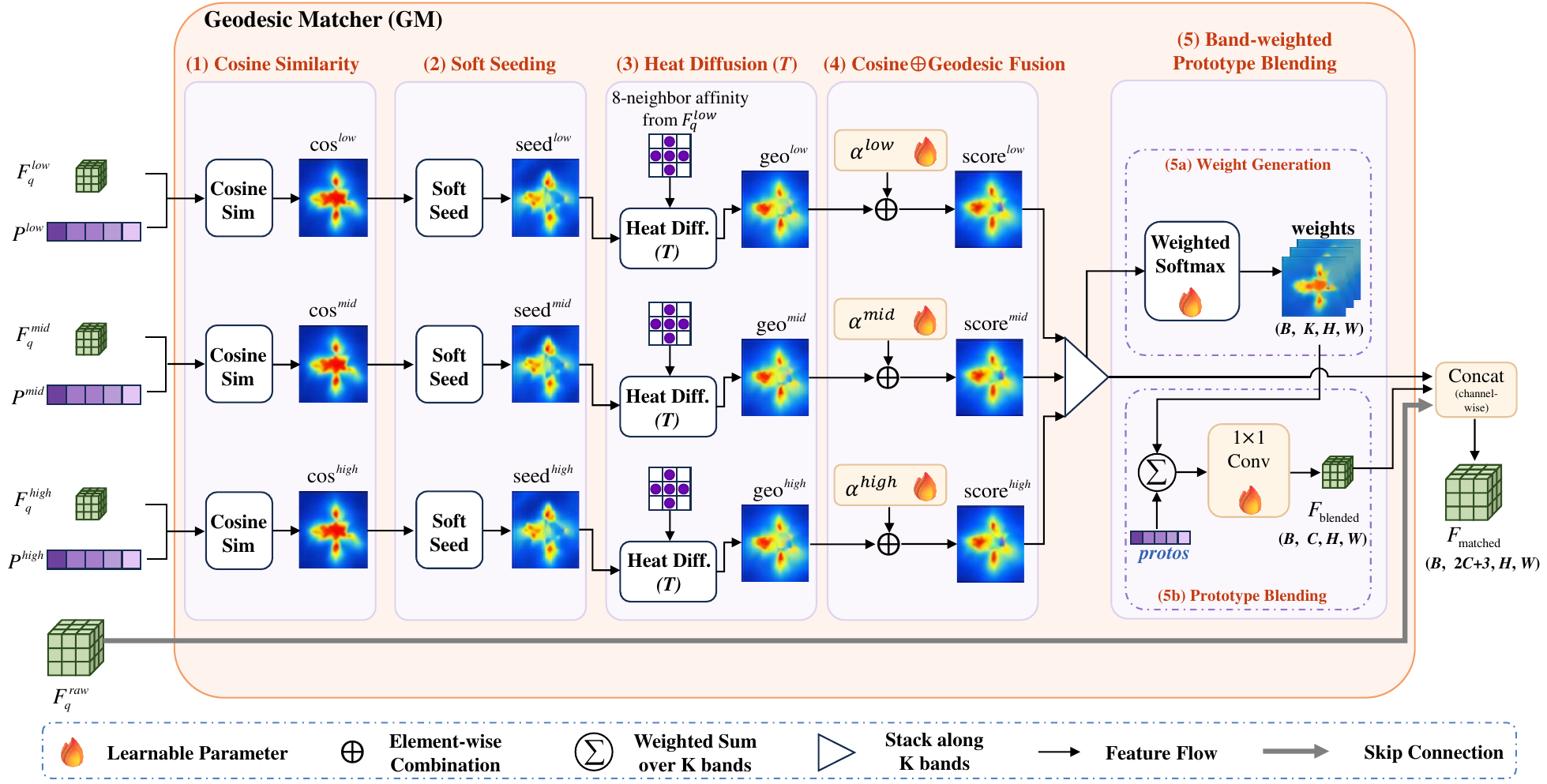}
\caption{Architecture of the Geodesic Matcher (GM). For each frequency band, GM (1) computes a cosine similarity $\cos^{(k)}$ between the query band feature and the band prototype, (2) converts it into a soft seed map, (3) refines it through $T$ heat-diffusion steps over an 8-neighbour feature affinity, and (4) fuses cosine and geodesic signals via a learnable gate $\alpha^{(k)}$. The per-band scores then (5) drive a pixel-wise softmax that blends the three prototypes into $F_{\mathrm{matched}}$.}
\label{fig:GM}
\end{figure*}

\subsection{Geodesic Matcher}
Existing FSMIS methods \cite{r10,r38,r12,lms-net} uniformly close the prototype--query matching loop with cosine similarity, which measures distances in the ambient Euclidean space and ignores the intrinsic feature manifold on which deep organ representations lie. This leads to broken intra-region connectivity in low-contrast areas and weakened inter-region separability between disconnected look-alike regions. We therefore propose a Geodesic Matcher (GM) built on the Heat Method, a one-shot differentiable approximation of geodesic distance via Varadhan's formula, which propagates the matching signal along a feature affinity graph through $T$ heat-diffusion steps and blends the three band prototypes per pixel.

GM receives $(F_q^{\mathrm{raw}}, bands_q, \mathcal{P})$ from the SPB and produces a Decoder-compatible tensor $F_{\mathrm{matched}} \in \mathbb{R}^{B \times (2C+3) \times h \times w}$. As shown in Fig.~\ref{fig:GM}, it consists of five stages: Cosine Similarity, Soft Seeding, Heat Diffusion, Cosine--Geodesic Fusion, and Band-Weighted Prototype Blending.

\subsubsection{Cosine Similarity}
For every band $k$, we compute
\begin{align} \label{eq:6}
\cos^{(k)}(i,j) = \frac{\langle F_q^{(k)}(i,j),\, P^{(k)} \rangle}{\| F_q^{(k)}(i,j) \| \, \| P^{(k)} \|},
\end{align}
which provides the same baseline matching signal used in mainstream FSMIS pipelines and serves as the source of geodesic refinement.

\subsubsection{Soft Seeding}
The Heat Method requires source pixels from which the heat is diffused. To keep this step differentiable, we use a soft scheme: with $\tau^{(k)}$ being the $q$-th quantile of $\cos^{(k)}$ ($q = 0.85$ by default, computed under stop-gradient),
\begin{align} \label{eq:7}
\mathrm{seed}^{(k)}(i,j) = \sigma\!\big(s\,(\cos^{(k)}(i,j) - \tau^{(k)})\big),
\end{align}
where $s = 20$ is a fixed sharpness scalar. Every pixel still contributes some heat, weighted by its cosine confidence.

\subsubsection{Heat Diffusion}
Varadhan's formula, $d_{\mathrm{geo}}^2(x, y) = \lim_{t \to 0^+} -4t \log u_t(x, y)$, links the geodesic distance to the heat kernel $u_t$. On a discrete 8-neighbour grid, one heat step amounts to an affinity-weighted local average, so iterating the discrete heat step $T$ times approximates the continuous heat kernel at a small time and remains fully differentiable. For each band, the affinity tensor is built from $F_q^{(k)}$ so that the geodesic respects the anatomical scale captured by that band. Let $\widehat{F}_q^{(k)}$ be the L2-normalised band feature and $(\delta y_n, \delta x_n)$ for $n = 1, \ldots, 8$ be the eight neighbour shifts. The affinity at site $(i,j)$ along shift $n$ is
\begin{equation} \label{eq:8}
\begin{aligned}
\mathcal{A}_n^{(k)}(i,j) =\ & \exp\!\Big(\!-\tfrac{1 - c_n^{(k)}(i,j)}{\sigma_a}\Big) \cdot \mathbf{1}_{\Omega}(i+\delta y_n, j+\delta x_n), \\[2pt]
c_n^{(k)}(i,j) =\ & \big\langle \widehat{F}_q^{(k)}(i,j),\, \widehat{F}_q^{(k)}(i+\delta y_n, j+\delta x_n) \big\rangle,
\end{aligned}
\end{equation}
where $\sigma_a = 0.5$ controls the connectivity sparsity and $\mathbf{1}_{\Omega}$ kills out-of-bound shifts. With $u^{(0,k)} = \mathrm{seed}^{(k)}$, one Jacobi-style step with self-loop weight one is then
\begin{equation} \label{eq:9}
\begin{aligned}
u^{(t+1,k)}(i,j) =\ & \frac{u^{(t,k)}(i,j) + \mathcal{H}^{(k,t)}(i,j)}{1 + \sum_{n=1}^{8} \mathcal{A}_n^{(k)}(i,j)}, \\
\mathcal{H}^{(k,t)}(i,j) =\ & \sum_{n=1}^{8} \mathcal{A}_n^{(k)}(i,j)\, u^{(t,k)}\!\big(\boldsymbol{p}_n(i,j)\big),
\end{aligned}
\end{equation}
where $\boldsymbol{p}_n(i,j) = (i + \delta y_n,\, j + \delta x_n)$ denotes the $n$-th neighbour location, and the unit self-loop in the denominator keeps the update well-defined when all neighbour affinities collapse to zero. After $T$ iterations we take $\mathrm{geo}^{(k)} = u^{(T,k)} \in [0,1]^{B \times 1 \times h \times w}$ as the geodesic reachability score. The full GM forward pass is summarised in Algorithm~\ref{alg:gm}.

\begin{algorithm}[t]
\caption{Geodesic Matcher (GM): one forward pass.}
\label{alg:gm}
\begin{algorithmic}[1]
\Require $F_q^{\mathrm{raw}}\!\in\!\mathbb{R}^{B \times C \times h \times w}$;\ $\{F_q^{(k)}\}_{k=1}^{K}$;\ $\{P^{(k)}\}_{k=1}^{K}$.
\Require Parameters: $\{\tilde\alpha^{(k)}\},\{\ell^{(k)}\},\sigma_a,s,q,T$.
\Ensure $F_{\mathrm{matched}}\!\in\!\mathbb{R}^{B \times (2C+3) \times h \times w}$.
\For{$k = 1$ \textbf{to} $K$} \Comment{Stage 1}
    \State $\cos^{(k)} \gets \mathrm{CosSim}(F_q^{(k)}, P^{(k)})$
\EndFor
\For{$k = 1$ \textbf{to} $K$} \Comment{Stages 2--3}
    \State $\tau^{(k)} \gets \mathrm{Quantile}(\cos^{(k)}, q)$
    \State $u \gets \sigma\!\big(s(\cos^{(k)} - \tau^{(k)})\big)$
    \State $\mathcal{A}^{(k)} \gets \mathrm{Aff}_{8\text{-nbr}}(F_q^{(k)}, \sigma_a)$
    \For{$t = 1$ \textbf{to} $T$}
        \State $u \gets \mathrm{Diffuse}(u, \mathcal{A}^{(k)})$
    \EndFor
    \State $\mathrm{geo}^{(k)} \gets u$
\EndFor
\For{$k = 1$ \textbf{to} $K$} \Comment{Stage 4}
    \State $\alpha^{(k)} \gets \sigma(\tilde\alpha^{(k)})$
    \State $\mathrm{score}^{(k)} \gets (1{-}\alpha^{(k)})\cos^{(k)} + \alpha^{(k)}\mathrm{geo}^{(k)}$
\EndFor
\State $\mathcal{S} \gets \mathrm{Stack}_{k=1}^{K}\!\big(\mathrm{score}^{(1)},\ldots,\mathrm{score}^{(K)}\big)$
\State $\mathbf{w} \gets \mathrm{softmax}_{k}\!\big(s\,\boldsymbol{\ell}\odot\mathcal{S}\big)$
\State $F_{\mathrm{blended}} \gets \mathrm{Conv}_{1\times1}\!\big(\textstyle\sum_{k=1}^{K}\mathbf{w}^{(k)}\!\otimes\! P^{(k)}\big)$
\State $F_{\mathrm{matched}} \gets \mathrm{concat}_{C}(F_q^{\mathrm{raw}}, F_{\mathrm{blended}}, \mathcal{S})$
\State \Return $F_{\mathrm{matched}}$
\end{algorithmic}
\end{algorithm}

\subsubsection{Cosine--Geodesic Fusion}
Each band carries a scalar gate $\alpha^{(k)} = \sigma(\tilde\alpha^{(k)})$, with $\tilde\alpha^{(k)} = 0$ at initialisation so that $\alpha^{(k)} = 0.5$. The fused band score is then
\begin{align} \label{eq:10}
\mathrm{score}^{(k)} = (1 - \alpha^{(k)})\cos^{(k)} + \alpha^{(k)}\,\mathrm{geo}^{(k)},
\end{align}
which lets gradients flow into both pathways from the very first training step. Stacking the scores yields $\mathcal{S} \in \mathbb{R}^{B \times K \times h \times w}$.

\subsubsection{Band-Weighted Prototype Blending}
Different anatomical regions are best characterised by different bands, so we route the three prototypes per pixel by softmaxing the scaled scores across bands,
\begin{align} \label{eq:11}
w^{(k)}(i,j) = \frac{\exp\!\big(s\,\ell^{(k)}\,\mathrm{score}^{(k)}(i,j)\big)}{\sum_{k'} \exp\!\big(s\,\ell^{(k')}\,\mathrm{score}^{(k')}(i,j)\big)},
\end{align}
where $\ell^{(k)}$ is a learnable per-band logit (initialised to one). The weights satisfy $\sum_k w^{(k)}(i,j) = 1$, and each pixel receives its own convex combination of band prototypes, refined by a $1{\times}1$ convolution,
\begin{align} \label{eq:12}
F_{\mathrm{blended}}(i,j) = \mathrm{Conv}_{1\times1}\!\Big(\sum_{k=1}^{K} w^{(k)}(i,j)\,P^{(k)}\Big).
\end{align}
This pixel-wise routing activates $P^{\mathrm{low}}$ in organ interiors, $P^{\mathrm{mid}}$ over parenchymal regions, and $P^{\mathrm{high}}$ along edges, recovering the disentanglement built by the SPB at the matching stage.

\subsubsection{Output Assembly}
The matcher concatenates the raw query feature, the blended prototype map, and the per-band score stack along the channel dimension,
\begin{align} \label{eq:13}
F_{\mathrm{matched}} = \big[F_q^{\mathrm{raw}};\, F_{\mathrm{blended}};\, \mathcal{S}\big] \in \mathbb{R}^{B \times (2C+3) \times h \times w},
\end{align}
where the skip path $F_q^{\mathrm{raw}}$ supplies a residual reference to the Decoder. As in the SPB, GM is invoked twice per episode under the foreground--background symmetric design with all parameters shared.

\subsection{Loss Function}
The training objective $\mathcal{L}_{\mathrm{total}}$ combines a primary segmentation loss, a boundary-aware loss, and a role-swapped alignment loss. The first two supervise the dual-decoder predictions on the query image. The third re-uses the SPB and GM modules with the support and query roles swapped, so that prototype matching remains symmetric under role inversion.

Let $\hat{Y}_q \in [0,1]^{B \times 2 \times H \times W}$ be the soft query prediction (softmax over the concatenated background and foreground decoder logits) and $Y_q \in \{0, 1, \mathrm{ignore}\}^{B \times H \times W}$ be the ground-truth label. The primary loss is a class-weighted negative log-likelihood,
\begin{align} \label{eq:14}
\mathcal{L}_{\mathrm{prim}} = -\frac{1}{|\Omega_v|}\sum_{(i,j) \in \Omega_v} w_{Y_q(i,j)}\,\log\hat{Y}_q\!\big(Y_q(i,j),\, i, j\big),
\end{align}
where $\Omega_v$ collects locations whose label is not the ignore index, and $w \in \mathbb{R}^2$ compensates for the foreground--background imbalance in medical scans. To sharpen the predicted contour, we further use a boundary loss,
\begin{align} \label{eq:15}
\mathcal{L}_{\mathrm{b}} = \frac{1}{|\Omega|}\!\sum_{(i,j)\in\Omega} \big\| \mathcal{B}_{\theta_0}(\hat{Y}_q^{\mathrm{fg}})(i,j) - \mathcal{B}_{\theta}(Y_q^{\mathrm{fg}})(i,j) \big\|_2^2,
\end{align}
where $\mathcal{B}_{\theta}(\cdot)$ extracts a soft boundary band of radius $\theta$ via differentiable morphology, with $\theta_0 = 3$ and $\theta = 5$ following.

For the alignment loss, the soft query prediction is binarised into a pseudo-mask $\widetilde{M}_q$, and SPB and GM are re-run with $\widetilde{M}_q$ in place of $M_s$ and $I_s$ as the new query, yielding $\hat{Y}_s$ supervised by the real support label $Y_s$,
\begin{align} \label{eq:16}
\mathcal{L}_{\mathrm{align}} = \mathcal{L}_{\mathrm{prim}}(\hat{Y}_s, Y_s) + \mathcal{L}_{\mathrm{b}}(\hat{Y}_s, Y_s).
\end{align}
For numerical stability in the cold-start regime, the foreground and background pathways always run in parallel, and any pseudo-mask with no positive pixels is replaced by a uniform mask. The overall objective is the unweighted sum of the three terms,
\begin{align} \label{eq:17}
\mathcal{L}_{\mathrm{total}} = \mathcal{L}_{\mathrm{prim}} + \mathcal{L}_{\mathrm{b}} + \mathcal{L}_{\mathrm{align}}.
\end{align}

The whole network, including the encoder, the SPB, the GM, and the two decoders, is trained end-to-end by minimising $\mathcal{L}_{\mathrm{total}}$.

\section{Experiments}
\subsection{Experiment Settings}
\subsubsection{Datasets}
We evaluate SGP-Net on three publicly available medical image segmentation benchmarks that span two anatomical regions and three imaging modalities, following the standard FSMIS evaluation protocol \cite{r5,r10}.
\begin{itemize}
\item \textbf{Abd-MRI (CHAOS-T2).} Released as Task 5 of the 2019 ISBI Combined Healthy Abdominal Organ Segmentation challenge \cite{r33}, this dataset contains 20 T2-SPIR abdominal MRI scans annotated for four abdominal organs: left kidney (LK), right kidney (RK), liver, and spleen.
\item \textbf{Abd-CT (SABS).} Drawn from the 2015 MICCAI Multi-Atlas Abdomen Labeling challenge \cite{r34}, this dataset comprises 30 abdominal CT scans with the same four organ classes as Abd-MRI, providing a direct cross-modality counterpart.
\item \textbf{CMR.} Released as part of the 2019 MICCAI Multi-sequence Cardiac MRI Segmentation challenge \cite{r35}, this dataset contains 35 cardiac MRI scans annotated for three structures: left-ventricle blood pool (LV-BP), left-ventricle myocardium (LV-MYO), and right ventricle (RV).
\end{itemize}

For consistency with prior work, all 3-D volumes are reformatted into 2-D axial slices and resized to $256\times256$.
\subsubsection{Experimental Settings}
We adopt the standard 1-way 1-shot episodic protocol and evaluate under two supervision settings introduced in \cite{r10,r11}, referred to as Setting 1 and Setting 2. In Setting 1, slices that contain test-class organs may still appear in the training set as long as the corresponding pixels are not annotated as the test class. Setting 2 is stricter: any slice containing the test class is removed entirely from training, ensuring that the model never sees the test category in any form during meta-training. Setting 2 is therefore a closer approximation of the truly unseen-class scenario, but is not applicable to CMR, where all three cardiac structures co-occur in nearly every slice. For each dataset, we adopt a 5-fold cross-validation scheme: in each fold, one organ is held out as the novel test class while the remaining organs serve as base classes for training.

\subsubsection{Implementation Details}
SGP-Net is implemented in PyTorch and trained on a single NVIDIA RTX 3090 GPU. Following the common practice of recent FSMIS works \cite{r12,r13}, the feature encoder uses a ResNet-101 backbone pretrained on part of MS-COCO \cite{r36}, with dilated convolutions in the last two stages so that the output stride is 8. Pseudo-masks for self-supervised meta-training are generated by the supervoxel clustering procedure of \cite{r5}.

The model is optimised end-to-end with SGD using an initial learning rate of $1\times 10^{-3}$, momentum 0.9, weight decay $5\times 10^{-4}$, and a step-decay schedule with $\gamma =0.9$. Training runs for 40k iterations with 10k iterations per epoch. The batch size is set to 1 following the episodic convention. The Spectral Prototype Bank uses $K=3$ frequency bands, with cut-off radii initialised to $(r_1,r_2)=(0.25,0.55)$ and learnt jointly with the rest of the network. The Geodesic Matcher uses $T=5$ heat-diffusion iterations and an affinity bandwidth $\sigma_a=0.5$, selected via the ablation study in Section~\ref{sec:ablation}. The seeding quantile is fixed to $q = 0.85$ across all experiments.

\subsubsection{Evaluation Metrics}
We use the Dice Similarity Coefficient (DSC) as the primary evaluation metric, defined for a predicted mask $\hat{Y}$ and ground truth $Y$ as:
\begin{align}
    \text{DSC}(\hat{Y}, Y) = \frac{2|\hat{Y} \cap Y|}{|\hat{Y}| + |Y|} \times 100\%.
\end{align}
To assess 2-D models on 3-D volumetric data, we follow the chunk-based evaluation protocol of \cite{r25}: each support and query scan is divided into a fixed number of chunks along the axial axis, and the median slice of each support chunk is used as the reference slice for all query slices in the corresponding chunk. The final per-class DSC is averaged over all five cross-validation folds, and the mean DSC over all classes within a dataset is reported as the overall performance indicator.

\subsection{Comparison with State-of-the-art Methods}

\begin{table*}[t]
\centering
\caption{Quantitative comparison in Dice score (\%) on the Abd-MRI and Abd-CT datasets under Setting~1 and Setting~2. Methods marked with \dag\ are reproduced under our unified protocol, others are quoted from the original papers. The best and second-best results are highlighted in \textbf{bold} and \underline{underlined}, respectively. ``--'' denotes results not reported in the original paper.}
\resizebox{\textwidth}{!}{
\begin{tabular}{c|c|c|ccccc|ccccc}
\toprule
\multirow{2}{*}{Setting} & \multirow{2}{*}{Method} & \multirow{2}{*}{Ref.} 
& \multicolumn{5}{c|}{Abd-MRI} 
& \multicolumn{5}{c}{Abd-CT} \\

& & 
& Spleen & Liver & LK & RK & Mean
& Spleen & Liver & LK & RK & Mean \\

\midrule

\multirow{12}{*}{1}
& PA-Net \cite{r8} & CVPR'19 & 40.58 & 50.40 & 30.99 & 32.19 & 38.53 & 36.04 & 49.55 & 20.67 & 21.19 & 32.86 \\
& SSL-ALPNet \cite{r5} & ECCV'20 & 72.18 & 76.10 & 81.92 & 85.18 & 78.84 & 70.96 & 78.29 & 72.36 & 71.81 & 73.35 \\
& ADNet \cite{r29} & MIA'22 & 72.29 & 82.11 & 73.86 & 85.80 & 78.51 & 63.48 & 77.24 & 72.13 & 79.06 & 72.97 \\
& Q-Net \cite{r15} & IntellSys'23 & 75.99 & 81.47 & 78.36 & 87.98 & 81.02 & -- & -- & -- & -- & -- \\
& PAMI \cite{r31} & TIM'24 & 76.37 & 82.59 & 81.83 & 88.73 & 82.38 & 72.38 & 81.32 & 76.52 & \textbf{80.57} & 77.69 \\
& DSPNet \dag \cite{r6} & MIA'25 & 68.57 & 76.61 & 82.40 & 85.69 & 78.32 & 66.77 & 70.68 & 79.42 & 73.93 & 72.70 \\
& AVT-ProNet \cite{r37} & TCSVT'25 & 73.42 & 83.91 & 81.70 & 90.06 & 82.27 & 80.49 & 83.51 & 79.68 & 78.70 & \underline{80.60} \\
& FAMNet \cite{famnet} & AAAI'25 & 72.31 & 80.60 & 75.70 & 85.49 & 78.52 & 77.81 & 83.32 & 77.74 & 76.56 & 78.86 \\
& UPRE-Net \dag \cite{r4} & TMI'25 & \underline{77.36} & \underline{83.92} & 84.22 & \underline{90.71} & \underline{84.05} & 79.13 & 81.10 & 80.76 & 77.83 & 79.71 \\
& DIFD \dag \cite{r13} & TMI'25 & 76.34 & 79.47 & \textbf{86.80} & 90.06 & 83.17 & 77.39 & 81.31 & \underline{82.00} & 76.88 & 79.40 \\
& MACCFA \dag \cite{maccfa} & PR'26 & 74.37 & 81.69 & 82.04 & 89.53 & 81.91 & \underline{79.70} & \textbf{84.16} & 79.86 & 72.46 & 79.04 \\
& \textbf{SGP-Net (Ours)} & \multicolumn{1}{c|}{---} & \textbf{77.44} & \textbf{84.10} & \underline{86.23} & \textbf{91.54} & \textbf{84.83} & \textbf{80.93} & \underline{83.66} & \textbf{82.39} & \underline{77.90} & \textbf{81.22} \\

\midrule

\multirow{12}{*}{2}
& PA-Net \cite{r8} & CVPR'19 & 50.90 & 42.26 & 53.45 & 38.64 & 46.33 & 29.59 & 38.42 & 32.34 & 17.37 & 29.43 \\
& SSL-ALPNet \cite{r5} & ECCV'20 & 67.02 & 73.05 & 73.63 & 78.39 & 73.02 & 60.25 & 73.65 & 63.34 & 54.82 & 63.02 \\
& ADNet \cite{r29} & MIA'22 & 59.44 & 77.03 & 59.64 & 56.68 & 63.20 & 50.97 & 70.63 & 48.41 & 40.52 & 52.63 \\
& Q-Net \cite{r15} & IntellSys'23 & 65.37 & 78.25 & 64.81 & 65.94 & 68.59 & -- & -- & -- & -- & -- \\
& PAMI \cite{r31} & TIM'24 & 75.80 & 81.09 & 74.51 & 86.73 & 79.53 & 71.95 & 74.13 & 72.36 & 67.54 & 71.49 \\
& DSPNet \dag \cite{r6} & MIA'25 & 67.01 & 77.84 & 75.19 & 82.33 & 75.59 & 65.30 & 66.56 & 69.72 & 61.89 & 65.87 \\
& AVT-ProNet \cite{r37} & TCSVT'25 & 68.43 & \underline{83.21} & 72.22 & 88.37 & 76.74 & 77.04 & \textbf{83.54} & 77.84 & 75.35 & 78.45 \\
& FAMNet \cite{famnet} & AAAI'25 & 67.50 & 80.53 & 73.04 & 83.74 & 76.20 & 73.64 & \underline{81.58} & 76.38 & 74.71 & 76.58 \\
& UPRE-Net \dag \cite{r4} & TMI'25 & \underline{72.58} & \textbf{83.42} & 81.07 & 87.11 & \underline{81.05} & \underline{77.22} & 78.60 & \underline{81.74} & \underline{80.31} & \underline{79.47} \\
& DIFD \dag \cite{r13} & TMI'25 & 71.67 & 81.40 & \underline{81.29} & \underline{88.65} & 80.75 & 76.38 & 80.73 & 80.94 & 77.10 & 78.79 \\
& MACCFA \dag \cite{maccfa} & PR'26 & 59.74 & 78.42 & 61.93 & 83.47 & 70.89 & 60.57 & 72.48 & 47.13 & 51.20 & 57.85 \\
& \textbf{SGP-Net (Ours)} & \multicolumn{1}{c|}{---} & \textbf{72.79} & 81.63 & \textbf{81.32} & \textbf{89.92} & \textbf{81.41} & \textbf{79.95} & 79.27 & \textbf{83.33} & \textbf{81.20} & \textbf{80.94} \\

\bottomrule
\end{tabular}
}
\label{tab:mri_ct_test}
\end{table*}

\begin{table}[t]
\centering
\caption{Quantitative comparison in Dice score (\%) on the CMR dataset under Setting~1. The best result is highlighted in \textbf{bold}, and the second-best is \underline{underlined}.}
\begin{tabular}{l|l|cccc}
\toprule
\multicolumn{1}{c|}{Method} & \multicolumn{1}{c|}{Ref.} & LV-BP & LV-MYO & RV & Mean \\
\midrule
PA-Net \cite{r8}        & ICCV'19       & 70.43 & 46.79 & 69.52 & 62.25 \\
SSL-ALPNet \cite{r5}    & ECCV'20       & 83.98 & 67.68 & 82.15 & 77.94 \\
ADNet \cite{r29}        & MIA'22        & 87.53 & 62.43 & 77.31 & 75.76 \\
Q-Net \cite{r15}        & IntelliSys'23 & \textbf{90.25} & 65.92 & 78.19 & 78.15 \\
PAMI \cite{r31}         & TIM'24        & 89.57 & 66.82 & 80.17 & 78.85 \\
DSPNet \dag \cite{r6}   & MIA'25        & 85.32 & 65.81 & 78.70 & 76.70 \\
AVT-ProNet \cite{r37}   & TCSVT'25      & 67.70 & \textbf{90.49} & 78.69 & 78.96 \\
FAMNet \cite{famnet}       & AAAI'25     & 86.80 & 62.64 & 72.71 & 74.05 \\
UPRE-Net \dag \cite{r4} & TMI'25        & 88.43 & 69.18 & \underline{83.66} & \underline{80.42} \\
DIFD \dag \cite{r13}    & TMI'25        & \underline{90.05} & 68.21 & 82.37 & 80.21 \\
MACCFA \dag \cite{maccfa}  & PR'26         & 89.38 & 66.49 & 80.16 & 78.68 \\
\textbf{SGP-Net (Ours)} & \multicolumn{1}{c|}{---} & 89.74 & \underline{69.83} & \textbf{83.91} & \textbf{81.16} \\
\bottomrule
\end{tabular}
\label{tab:cmr_test}
\end{table}

We compare SGP-Net with eleven representative FSMIS methods, including PA-Net~\cite{r8}, SSL-ALPNet~\cite{r5}, ADNet~\cite{r29}, Q-Net~\cite{r15}, PAMI~\cite{r31}, DSPNet~\cite{r6}, AVT-ProNet~\cite{r37}, FAMNet~\cite{famnet}, UPRE-Net~\cite{r4}, DIFD~\cite{r13} and MACCFA~\cite{maccfa}. Methods marked with \dag\ are reproduced from their official codebases under our unified evaluation protocol, and the rest are quoted from the respective papers. Quantitative results on Abd-MRI and Abd-CT are summarised in Table~\ref{tab:mri_ct_test}, and results on CMR in Table~\ref{tab:cmr_test}.

\subsubsection{Quantitative results}
SGP-Net achieves the best mean Dice on all three datasets under both supervision settings. On Abd-MRI under Setting~1, SGP-Net obtains 84.83\%, exceeding the second-best competitor UPRE-Net by 0.78\%, with substantial gains on the liver and the right kidney where boundaries are most prone to inter-organ leakage. A similar trend is observed on Abd-CT, where SGP-Net reaches 81.22\% and outperforms AVT-ProNet by 0.62\%. On CMR, SGP-Net leads on the mean Dice and obtains the highest score of 83.91\% on the right ventricle, whose thin and often discontinuous wall is hard to capture by plain cosine matching. These per-organ improvements are consistent with our design: the spectral prototype bank dedicates a separate prototype to high-frequency boundary cues, while the geodesic matcher restores connectivity within organ interiors that have weak contrast against the surrounding tissue.

\subsubsection{Robustness under Setting 2}
Setting 2 removes every slice containing the test class from training and is therefore a more faithful evaluation of unseen-class generalisation. Most competitors suffer noticeable degradation under this regime. For instance, MACCFA drops from 79.04\% to 57.85\% on Abd-CT, and AVT-ProNet drops from 82.27\% to 76.74\% on Abd-MRI. SGP-Net is markedly more stable, losing only 0.28\% on Abd-CT and remaining the top performer on both datasets, with a 1.47\% advantage over UPRE-Net on Abd-CT and 0.36\% on Abd-MRI. The reason is that geodesic matching propagates similarity along the feature manifold rather than through ambient Euclidean space. When the test distribution shifts away from the training one, cosine similarity is easily attracted to off-manifold but visually similar regions, while heat diffusion remains anchored to the connected component containing the prototype and is thus less affected by distribution drift.

\subsubsection{Visualisation comparison}
\begin{figure}[t]
\centering
\includegraphics[width=\linewidth]{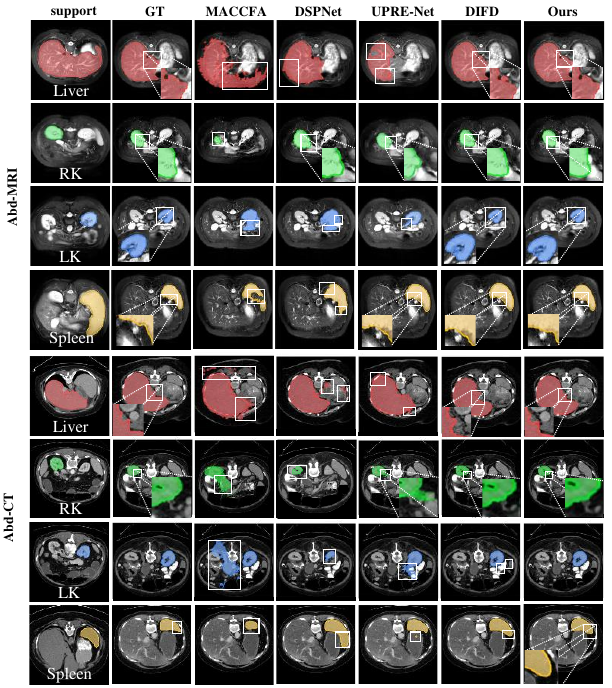}
\caption{Qualitative comparison on the Abd-MRI (top) and Abd-CT (bottom) datasets. From left to right: support image, ground truth, MACCFA, DSPNet, UPRE-Net, DIFD, and SGP-Net (Ours). Zoomed patches highlight boundary regions. Each row corresponds to a different organ class.}
\label{fig:vis_abd}
\end{figure}

\begin{figure}[t]
\centering
\includegraphics[width=\linewidth]{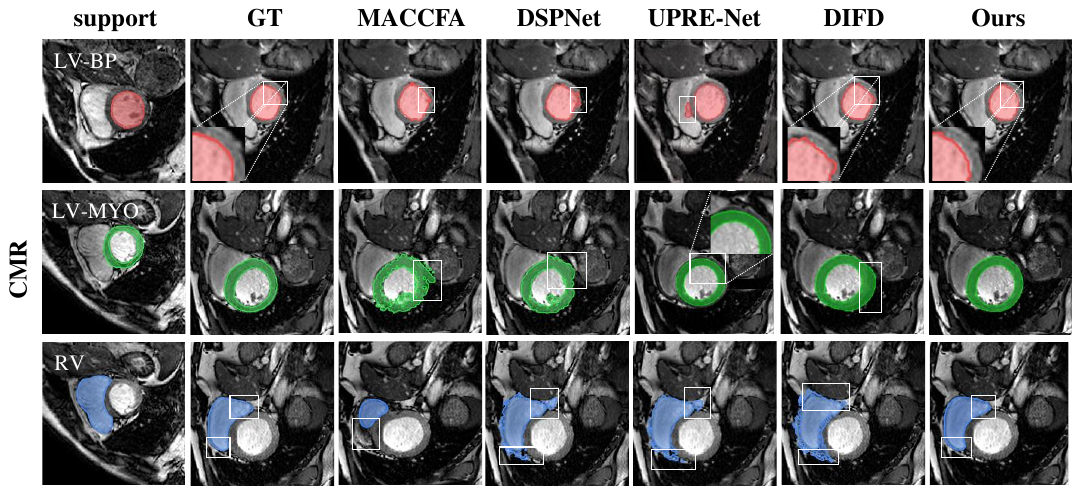}
\caption{Qualitative comparison on the CMR dataset. From left to right: support image, ground truth, MACCFA, DSPNet, UPRE-Net, DIFD, and SGP-Net (Ours). Rows correspond to LV-BP, LV-MYO, and RV, respectively.}
\label{fig:vis_cmr}
\end{figure}

Fig.~\ref{fig:vis_abd} and Fig.~\ref{fig:vis_cmr} present qualitative segmentation results on representative slices from the three datasets, with zoomed patches highlighting boundary regions. MACCFA and DSPNet tend to produce fragmented masks inside low-contrast organs such as the liver and the spleen, indicating that their cosine-based matching is broken by intensity inhomogeneity. DIFD and UPRE-Net leak across organ boundaries on the kidneys in Abd-CT, where neighbouring tissues share similar appearance with the target. On CMR, competing methods produce ragged contours on the myocardium and the right ventricle. SGP-Net delivers masks that are both more complete inside the organ and tighter along the boundary. This behaviour reflects the band-weighted prototype blending in the geodesic matcher, in which low-frequency prototypes activate organ interiors, mid-frequency prototypes refine parenchymal regions, and high-frequency prototypes sharpen contours, with all three routed by a connectivity-aware similarity map.

\subsection{Ablation Study}
\label{sec:ablation}    
We conduct extensive ablations on Abd-MRI under Setting~1 to verify the contribution of each component and the choice of key hyper-parameters. All ablations follow the main training protocol, and we report the mean Dice across five-fold cross-validation.

\subsubsection{Effect of the Spectral--Geodesic Prototype Module}
Table~\ref{tab:ablation_module} dissects the contribution of SPB and GM. The baseline replaces both with a single masked-average prototype matched by cosine similarity, as used in PANet~\cite{r8} and SSL-ALPNet~\cite{r5}. Adding SPB alone lifts the mean Dice from 79.42\% to 82.18\%, confirming that frequency-band decomposition yields more discriminative class representations than a single entangled prototype. Replacing cosine matching with GM brings a separate gain of 3.11\%, indicating that geodesic similarity captures intra-organ connectivity missed by cosine matching. Combining the two yields a total gain of 5.41\%, close to the sum of their individual gains, which confirms that SPB and GM address two independent failure modes.

\begin{table}[!t]
\centering
\caption{Ablation of the Spectral Prototype Bank (SPB) and the Geodesic Matcher (GM) on Abd-MRI under Setting~1.}
\label{tab:ablation_module}
\begin{tabular}{cc|cccc|c}
\toprule
SPB & GM & Spleen & Liver & LK & RK & Mean \\
\midrule
        &        & 73.18 & 79.65 & 80.27 & 84.57 & 79.42 \\
\checkmark &        & 75.94 & 82.36 & 82.71 & 87.69 & 82.18 \\
        & \checkmark & 76.13 & 82.84 & 83.05 & 88.10 & 82.53 \\
\checkmark & \checkmark & \textbf{77.44} & \textbf{84.10} & \textbf{86.23} & \textbf{91.54} & \textbf{84.83} \\
\bottomrule
\end{tabular}
\end{table}

\subsubsection{Number of frequency bands $K$ and heat-diffusion steps $T$}
\begin{figure}[!t]
\centering
\includegraphics[width=\linewidth]{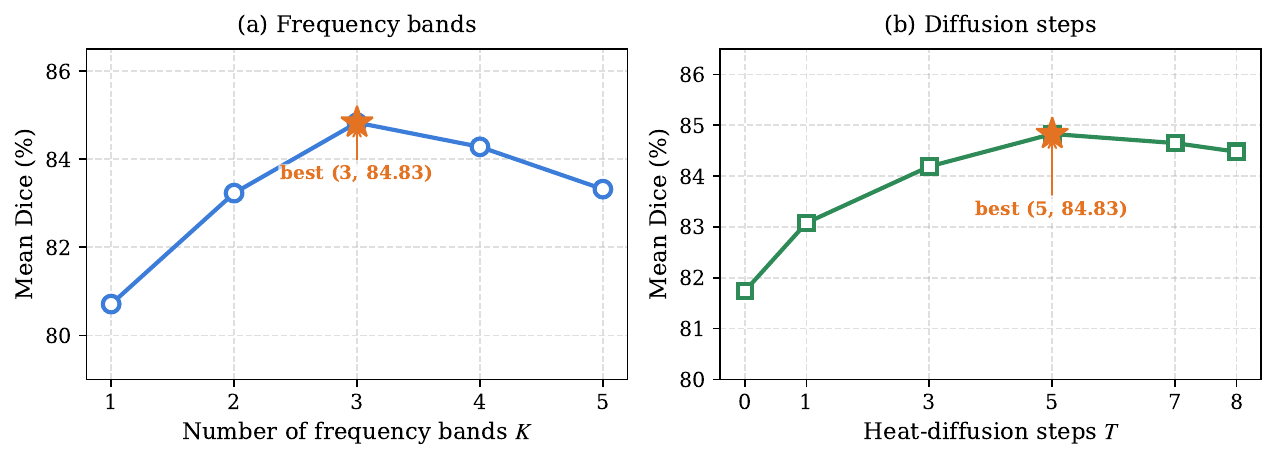}
\caption{Sensitivity of SGP-Net to $K$ (left) and $T$ (right). Both curves peak at the chosen settings ($K=3$, $T=5$) and remain stable in their neighbourhoods.}
\label{fig:k_curve}
\end{figure}

Tables~\ref{tab:ablation_bands} and \ref{tab:ablation_steps} report the effect of varying $K$ and $T$, jointly visualised in Fig.~\ref{fig:k_curve}. Setting $K=1$ degenerates SPB into a single full-spectrum prototype and produces only marginal gains; increasing $K$ to 3 lifts the mean Dice to 84.83\% by allowing silhouette, texture and boundary cues to be carried separately, while $K=4$ or $5$ brings diminishing returns due to overlapping band content. For $T$, performance climbs from 81.74\% at $T=0$ (GM reduces to a soft-seeded cosine map) and saturates around $T=5$, indicating that five iterations are sufficient for heat to traverse a typical organ-scale region. We therefore fix $K=3$ and $T=5$.

\begin{table}[!t]
\centering
\caption{Effect of the number of frequency bands $K$ on Abd-MRI under Setting~1.}
\label{tab:ablation_bands}
\begin{tabular}{c|cccc|c}
\toprule
$K$ & Spleen & Liver & LK & RK & Mean \\
\midrule
1 & 74.62 & 80.81 & 81.47 & 85.93 & 80.71 \\
2 & 76.18 & 82.95 & 84.36 & 89.42 & 83.23 \\
\textbf{3} & \textbf{77.44} & \textbf{84.10} & \textbf{86.23} & \textbf{91.54} & \textbf{84.83} \\
4 & 76.91 & 83.62 & 85.74 & 90.86 & 84.28 \\
5 & 76.05 & 82.71 & 84.59 & 89.93 & 83.32 \\
\bottomrule
\end{tabular}
\end{table}

\begin{table}[!t]
\centering
\caption{Effect of the number of heat-diffusion steps $T$ on Abd-MRI under Setting~1.}
\label{tab:ablation_steps}
\begin{tabular}{c|cccc|c}
\toprule
$T$ & Spleen & Liver & LK & RK & Mean \\
\midrule
0 & 75.36 & 81.79 & 82.04 & 87.78 & 81.74 \\
1 & 76.42 & 82.86 & 83.91 & 89.13 & 83.08 \\
3 & 77.05 & 83.74 & 85.36 & 90.62 & 84.19 \\
\textbf{5} & \textbf{77.44} & \textbf{84.10} & \textbf{86.23} & \textbf{91.54} & \textbf{84.83} \\
7 & 77.31 & 83.97 & 86.05 & 91.27 & 84.65 \\
8 & 77.18 & 83.82 & 85.81 & 91.12 & 84.48 \\
\bottomrule
\end{tabular}
\end{table}

\subsubsection{Visualisation of spectral disentanglement}
\begin{figure}[!t]
\centering
\includegraphics[width=\linewidth]{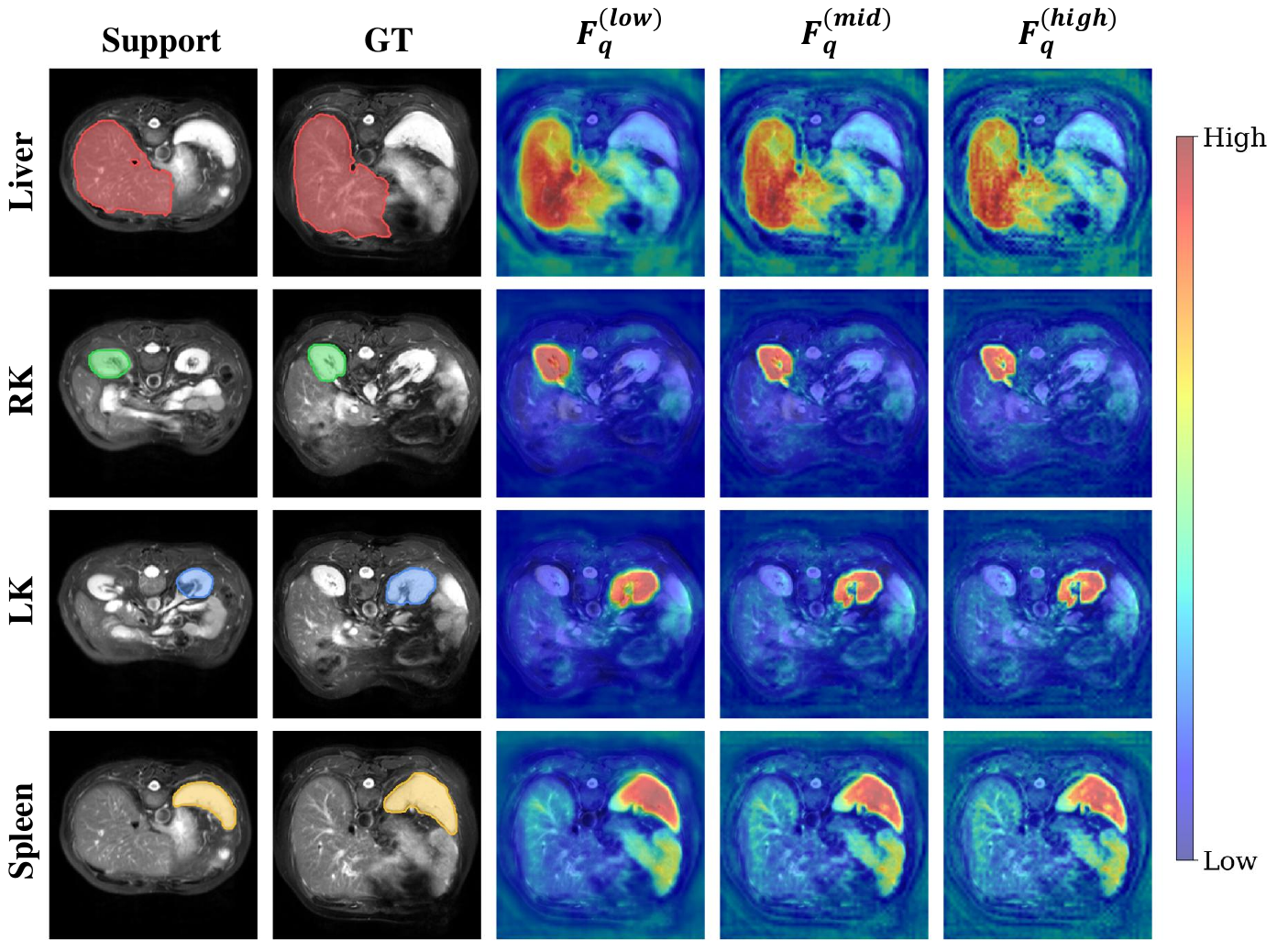}
\caption{Per-band activation maps produced by the Spectral Prototype Bank on Abd-MRI. From left to right: support image with mask, query image with ground truth, and the cosine activation maps of the low-, mid- and high-frequency band prototypes overlaid on the query slice. Warmer colours denote higher activation.}
\label{fig:band_act}
\end{figure}

The module ablation in Table~\ref{tab:ablation_module} confirms that the SPB brings substantial gains, but does not directly explain \emph{what} each frequency band captures. To probe this, we visualise the per-band cosine activation maps on representative Abd-MRI slices in Fig.~\ref{fig:band_act}. For every organ, the low-frequency map fires uniformly across the entire organ region and recovers the global silhouette, the mid-frequency map preserves the silhouette while progressively highlighting internal parenchymal structure, and the high-frequency map further concentrates the response near the organ contour. This division of labour is consistent across all four organs and matches the design intent of the SPB: the three radial bands carry complementary cues that, when blended pixel-wise by the Geodesic Matcher, yield matching scores that are simultaneously cue-aware and spatially complete.

\subsubsection{Convergence of the learnable cut-off radii}
\begin{figure}[!t]
\centering
\includegraphics[width=0.85\linewidth]{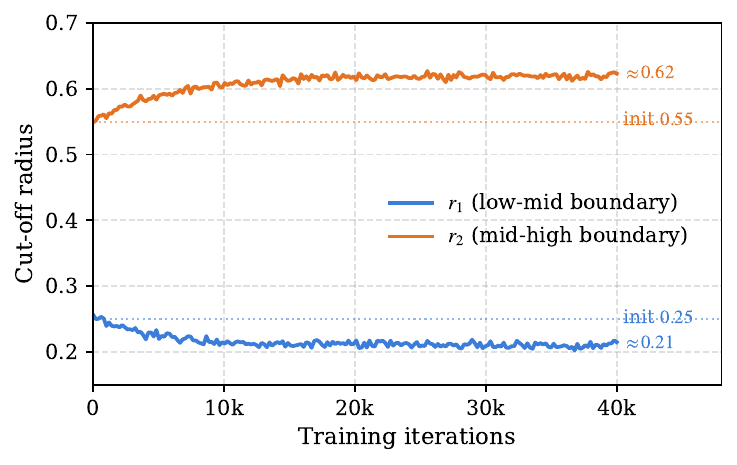}
\caption{Trajectory of the learnable cut-off radii $r_1$ and $r_2$ during training on Abd-MRI. Both radii converge within 20k iterations, with the high-frequency band ($r > r_2$) gradually widening to better capture boundary cues.}
\label{fig:radii_curve}
\end{figure}

The two cut-off radii $r_1$ and $r_2$ that partition the Fourier spectrum are jointly learned with the rest of the network, rather than hand-set. Fig.~\ref{fig:radii_curve} traces their values throughout training. Both radii start from their initial settings of $r_1 = 0.25$ and $r_2 = 0.55$ and converge smoothly to $r_1 \approx 0.21$ and $r_2 \approx 0.62$ within roughly 20k iterations. The widening gap reflects an interpretable optimisation outcome: the model assigns a relatively narrower low-frequency band for compact silhouette information and a broader high-frequency band for boundary refinement, which is consistent with the per-band division of labour described in Section~\ref{subsec:spb}.

\subsubsection{Stability across cross-validation folds}
\begin{figure}[!t]
\centering
\includegraphics[width=\linewidth]{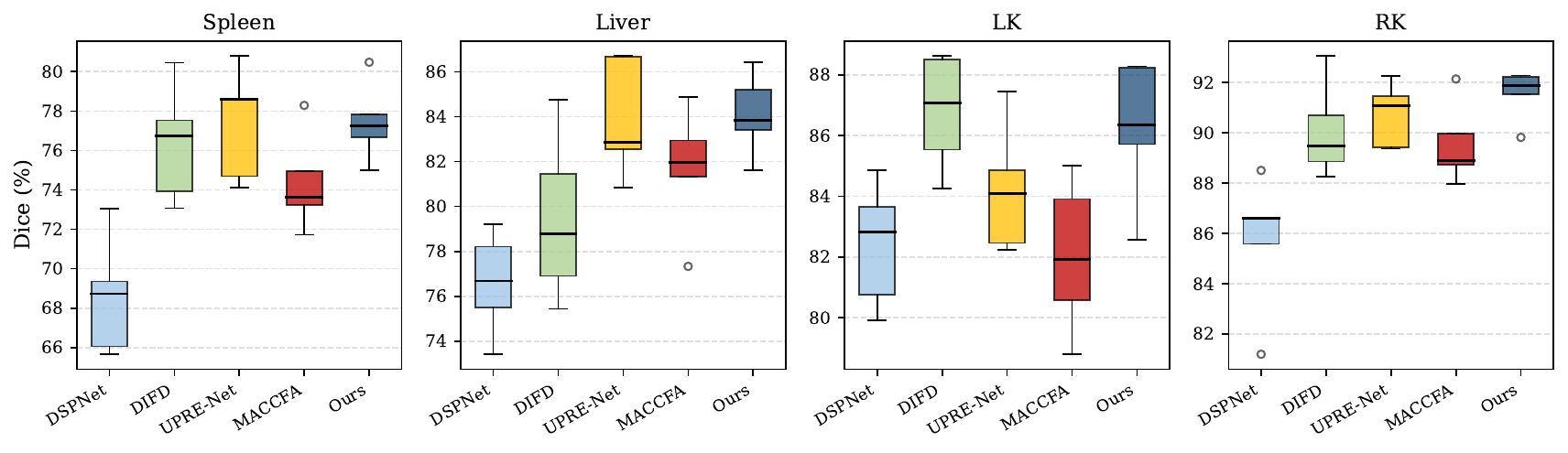}
\caption{Per-fold Dice distributions on the four Abd-MRI organs across five-fold cross-validation. SGP-Net achieves the highest median Dice and the smallest interquartile range on every organ.}
\label{fig:boxplot}
\end{figure}

Few-shot segmentation is sensitive to which support--query split is sampled, and a method that averages well but exhibits large fold-to-fold variance is undesirable in clinical practice. Fig.~\ref{fig:boxplot} reports the per-fold Dice distributions on the four Abd-MRI organs for SGP-Net and four competitive baselines. SGP-Net attains the highest or near-highest median across all four organs and produces noticeably tighter distributions, especially on the spleen and the left kidney, where prior methods exhibit long tails caused by support--query mismatch. The reduced variance can be traced back to geodesic matching: when a particular support slice is unrepresentative of the query distribution, cosine similarity propagates the mismatch directly into the prediction, whereas heat diffusion smooths the matching signal along the manifold and absorbs part of the noise.

\subsubsection{Robustness across supervision settings}
\begin{figure}[!t]
\centering
\includegraphics[width=\linewidth]{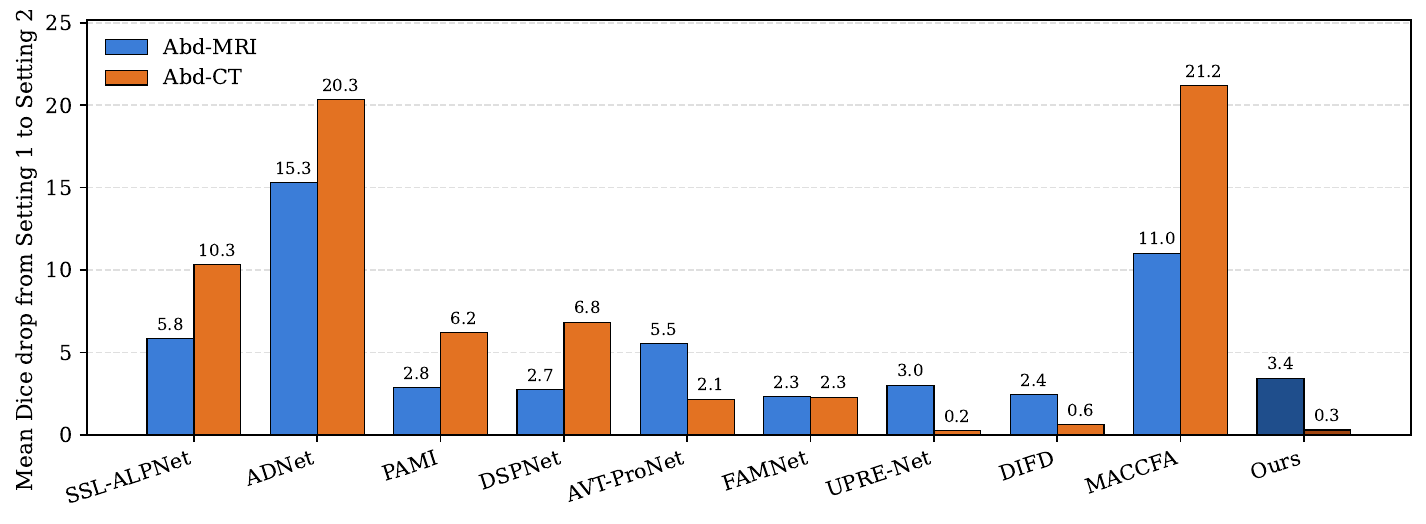}
\caption{Performance drop in mean Dice (\%) from Setting~1 to Setting~2 on Abd-MRI and Abd-CT. SGP-Net suffers only a marginal drop on both datasets while remaining the top performer under Setting~2.}
\label{fig:setting_drop}
\end{figure}

Setting~2 is markedly stricter than Setting~1, as it removes every slice containing the test class from training. Fig.~\ref{fig:setting_drop} compares the per-method drop on Abd-MRI and Abd-CT. While strong baselines such as MACCFA degrade by more than 20\% on Abd-CT, SGP-Net suffers only a marginal drop of 0.28\% on Abd-CT and 3.42\% on Abd-MRI, and remains the top performer on both datasets under Setting~2. We attribute this stability to the topology-aware nature of geodesic matching: cosine similarity is easily attracted to off-manifold but visually similar regions when the test distribution diverges from the training one, whereas heat diffusion is anchored to the connected component containing the prototype and is therefore more resilient to distribution shift.

\subsubsection{Computational complexity}
We further compare SGP-Net with five representative FSMIS methods in terms of the number of parameters, the number of floating-point operations (FLOPs) for a single 1-way 1-shot episode at input resolution $256\times 256$, and the inference speed measured on an NVIDIA RTX 3090 GPU. The results are summarised in Table~\ref{tab:complexity}. Recent prototype-based methods are significantly heavier than the early PA-Net baseline due to their richer prototype generation and matching modules. SGP-Net introduces only modest extra computation on top of strong recent competitors: it has fewer parameters than UPRE-Net and DIFD and runs at a comparable speed to DSPNet, so the overhead of the Spectral Prototype Bank and the Geodesic Matcher is well controlled.

\begin{table}[!t]
\centering
\caption{Comparison of model complexity on Abd-MRI under Setting~1. All methods are evaluated on a single NVIDIA RTX 3090 GPU.}
\label{tab:complexity}
\begin{tabular}{l|ccc|c}
\toprule
Method & Params (M) & FLOPs (G) & FPS & Mean Dice (\%) \\
\midrule
PA-Net~\cite{r8}        & \textbf{14.71} & \textbf{50.96} & \textbf{36.4} & 38.53 \\
SSL-ALPNet~\cite{r5}    & 38.97 & 84.13 & 28.7 & 78.84 \\
DSPNet~\cite{r6}        & 47.62 & 173.41 & 18.2 & 78.32 \\
DIFD~\cite{r13}         & 53.18 & 281.74 & 13.8 & 83.17 \\
UPRE-Net~\cite{r4}      & 54.06 & 285.92 & 13.1 & 84.05 \\
\textbf{SGP-Net (Ours)} & 51.82 & 279.63 & 14.6 & \textbf{84.83} \\
\bottomrule
\end{tabular}
\end{table}

\FloatBarrier



\section{Conclusion}

In this work, we presented SGP-Net, a few-shot medical image segmentation framework that addresses two limitations of standard prototype pipelines: the reliance on a single spatial-domain vector and the use of ambient Euclidean space for similarity measurement. At its core, the proposed Spectral--Geodesic Prototype Module pairs a Spectral Prototype Bank, which decomposes features into multi-frequency bands via learnable radial Fourier filters, with a Geodesic Matcher, which computes similarity on the feature manifold through a differentiable heat-diffusion approximation. The two components are coupled by a learnable gate and a band-weighted blending head, and operate symmetrically on the foreground and background. Experiments on Abd-MRI, Abd-CT, and CMR show that SGP-Net consistently outperforms state-of-the-art methods, and is notably robust against degradation when the test class is fully unseen during training. Ablation studies confirm the complementarity of the two modules, the stability of the selected hyper-parameters, and the interpretability of the learned cut-off radii.

In future work, we plan to extend the spectral--geodesic formulation beyond the current 1-way 1-shot intra-modality protocol to multi-way, multi-shot, and cross-modality scenarios, and to replace the fixed heat-diffusion step count $T$ with an adaptive scheme based on local feature affinity for better efficiency on volumetric data.

\FloatBarrier

\bibliographystyle{IEEEtran}
\bibliography{SGP-Net}

\begin{IEEEbiography}[{\includegraphics[width=1in,height=1.25in,clip,keepaspectratio]{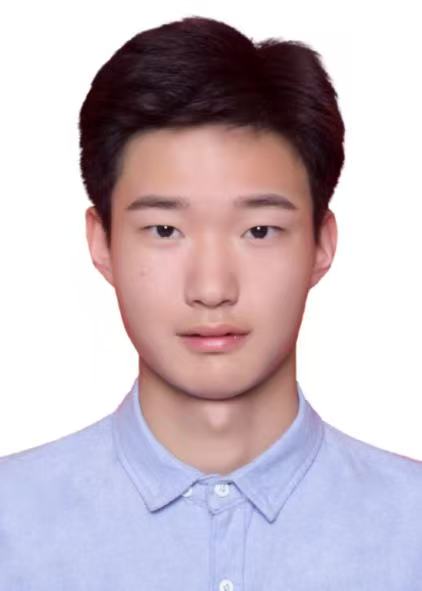}}]
{Penghao Jia} received the M.E. degree in computer technology from Chongqing University of Technology, China. He is currently pursuing the Ph.D. degree at Chongqing University. His research interests include few-shot learning and medical image segmentation.
\end{IEEEbiography}

\begin{IEEEbiography}[{\includegraphics[width=1in,height=1.25in,clip,keepaspectratio]{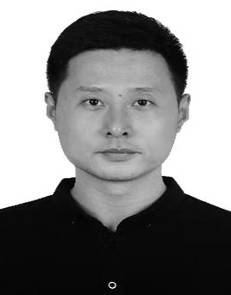}}]
{Zhiyong Huang} received the B.S. degree in electrical engineering from Chongqing University, Chongqing, China, in 2001, the M.S. degree in electronic engineering from Chongqing University, in 2004, and the Ph.D. degree in electronic engineering from Chongqing University, in 2010. He is
currently a Professor with the School of Microelectronics and Communication Engineering, Chongqing University. He has published a number of papers in renowned journals, including the IEEE TRANSACTIONS ON CIRCUITS AND SYSTEMS FOR VIDEO TECHNOLOGY, IEEE TRANSACTIONS ON INTELLIGENT TRANSPORTATION SYSTEMS, IEEE JOURNAL OF BIOMEDICAL AND HEALTH INFORMATICS, IEEE TRANSACTIONS ON GEOSCIENCE AND REMOTE SENSING, and IEEE SIGNAL PROCESSING LETTERS, among others. His research interests include medical artificial intelligence, data analysis, and machine learning.
\end{IEEEbiography}

\begin{IEEEbiography}[{\includegraphics[width=1in,height=1.25in,clip,keepaspectratio]{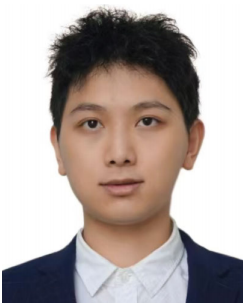}}]
{Mingyang Hou} is currently pursuing the Ph.D. degree with the School of Microelectronics and
Communication Engineering, Chongqing University. He has published several articles in reputable journals, including IEEE Transactions on Circuits and Systems for Video Technology, IEEE Transactions on Geoscience and Remote Sensing, and Computers in Biology and Medicine. His research interests include deep learning, image super-resolution, and medical image processing.
\end{IEEEbiography}

\begin{IEEEbiography}[{\includegraphics[width=1in,height=1.25in,clip,keepaspectratio]{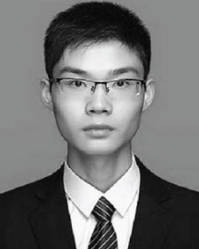}}]
{Zhi Yu} received the B.S. and M.S. degrees from the School of Microelectronics and Communication Engineering, Chongqing University, China, in 2017 and 2020, respectively, where he is currently pursuing the Ph.D. degree. He has published some articles in the refereed journals, such as IEEE Transactions on Intelligent Transportation Systems, Information Processing and Management, and Neural Computing and Applications. His research interests include the IoT and object re-identification.
\end{IEEEbiography}

\begin{IEEEbiography}[{\includegraphics[width=1in,height=1.25in,clip,keepaspectratio]{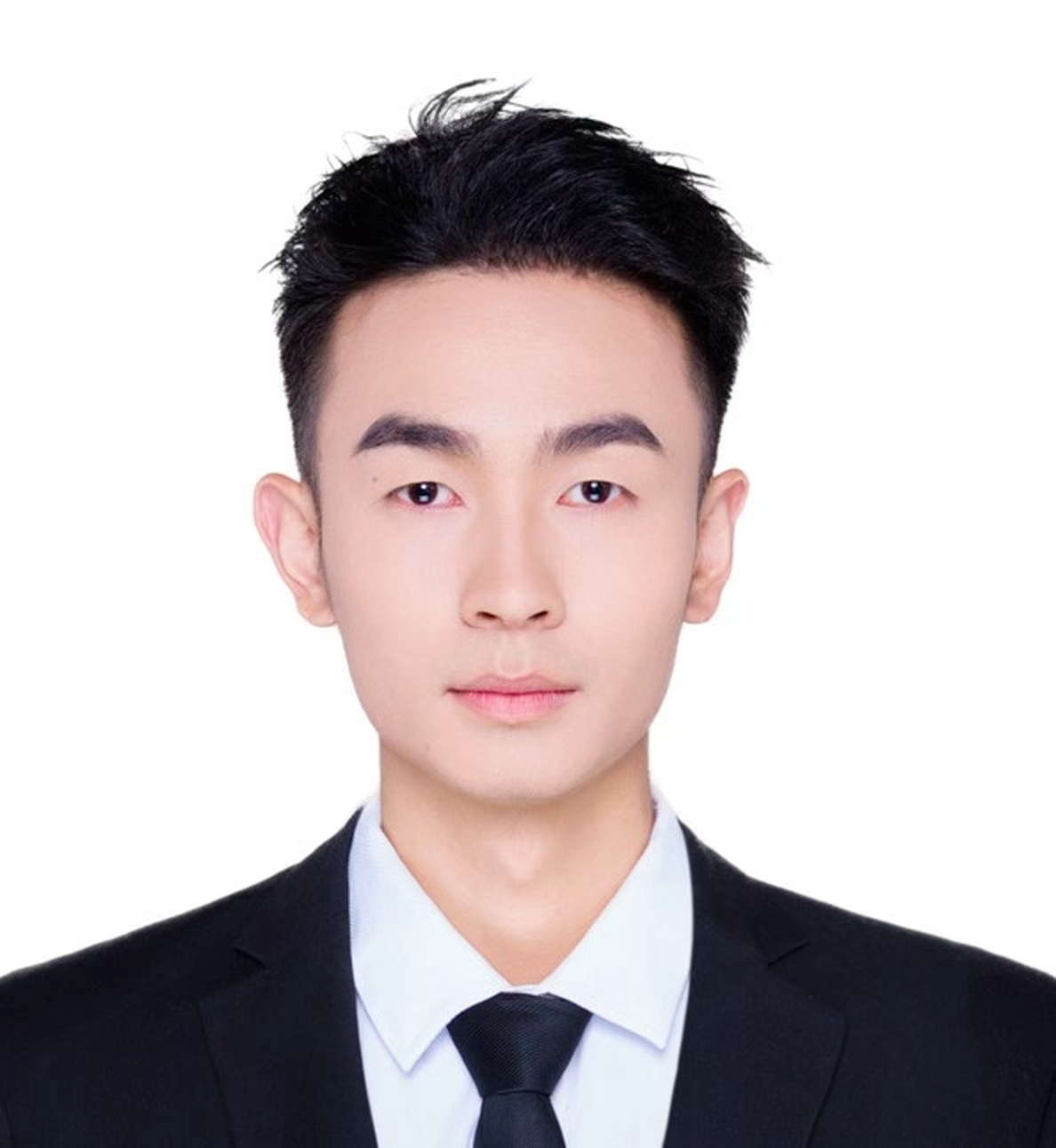}}]
{Shuai Miao} received a B.S. degree from Chongqing University, Chongqing, China, in 2020. He is pursuing a Ph.D. degree from the School of Microelectronics and Communication Engineering, Chongging University, Chongqing, China. His research interests include deep learning.
\end{IEEEbiography}

\begin{IEEEbiography}[{\includegraphics[width=1in,height=1.25in,clip,keepaspectratio]{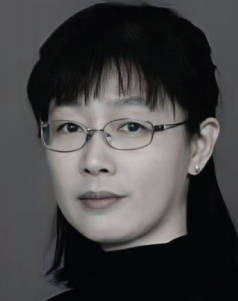}}]
{Jiahong Wang} is currently the Deputy Director of the Health Medicine Ward, Chinese People’s Liberation Army General Hospital. She is dedicated to research in pre-hospital and outpatient management and sports intervention in the field of health medicine. She participated in the research activities held under the “Capital Residents Cholesterol Education Program.” She won China Medical Science and Technology Award and undertook the 12th Five-Year National Support Program “Screening and Early Intervention of Metabolic Diseases.”
\end{IEEEbiography}

\begin{IEEEbiography}[{\includegraphics[width=1in,height=1.25in,clip,keepaspectratio]{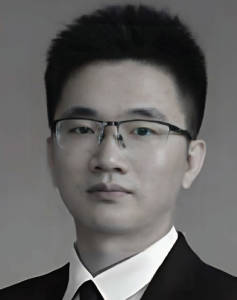}}]
{Yan Yan} (Member IEEE) received the B.Eng. and M.Sc. degrees in instrumentation engineering from Harbin Institute of Technology in 2010 and 2012, respectively, and the Ph.D. degree in computer science from the University of Chinese Academy of Sciences in 2020. From 2012 to 2014, he was a Research Assistant with Shenzhen Institutes of Advanced Technology, Chinese Academy of Sciences (SIAT-CAS), China. Additionally, he was an Honorary Research Assistant with the Department of Computer Science, University of Liverpool, from 2017 to 2018, under the mentorship of Prof. Y. Goulermas. He currently holds the position of an Assistant Researcher with SIAT-CAS and the Center of Scientific and Technological Service, Wenzhou Institute of Technology. His research interests include biomedical signal processing, pattern recognition, nonlinear dynamical systems, and topological data analysis.
\end{IEEEbiography}
\vfill
\end{document}